# PMLB v1.0: An open-source dataset collection for benchmarking machine learning methods

JOSEPH D. ROMANO*, University of Pennsylvania, USA and University of Pennsylvania, USA

TRANG T. LE*, University of Pennsylvania, USA

WILLIAM LA CAVA*, University of Pennsylvania, USA

JOHN T. GREGG, University of Pennsylvania, USA

DANIEL J. GOLDBERG, Washington University in St. Louis, USA

NATASHA L. RAY, Princeton Day School, USA

PRANEEL CHAKRABORTY, University of Pennsylvania, USA and University of Pennsylvania, USA

DANIEL HIMMELSTEIN, Washington University in St. Louis, USA and University of Pennsylvania, USA

WEIXUAN FU, University of Pennsylvania, USA

JASON H. MOORE†, University of Pennsylvania, USA

**Motivation:** Novel machine learning and statistical modeling studies rely on standardized comparisons to existing methods using well-studied benchmark datasets. Few tools exist that provide rapid access to many of these datasets through a standardized, user-friendly interface that integrates well with popular data science workflows.

**Results:** This release of PMLB provides the largest collection of diverse, public benchmark datasets for evaluating new machine learning and data science methods aggregated in one location. v1.0 introduces a number of critical improvements developed following discussions with the open-source community.

**Availability:** PMLB is available at https://github.com/EpistasisLab/pmlb. Python and R interfaces for PMLB can be installed through the Python Package Index and Comprehensive R Archive Network, respectively.

## 1 INTRODUCTION

Benchmarking is a standard practice for evaluating the strengths and weaknesses of machine learning (ML) algorithms with regard to different problem characteristics. Benchmarking involves assessing the performance of specific ML models—namely, how well they perform on a group of well-studied benchmark datasets [1, 4]. Although these benchmark datasets are plentiful, they are often difficult to access, challenging to integrate with analyses of other datasets, and prone to myriad data quality issues. PMLB (Penn Machine Learning Benchmarks) is a large, curated repository of open source benchmark datasets that aims to solve these issues.

*These authors contributed equally.
†Corresponding author.

Authors' addresses: Joseph D. Romano, University of Pennsylvania, Institute for Biomedical Informatics, Philadelphia, Pennsylvania, USA, University of Pennsylvania, Center of Excellence in Environmental Toxicology, Philadelphia, Pennsylvania, USA; Trang T. Le, University of Pennsylvania, Institute for Biomedical Informatics, Philadelphia, Pennsylvania, USA; William La Cava, University of Pennsylvania, Institute for Biomedical Informatics, Philadelphia, Pennsylvania, USA; John T. Gregg, University of Pennsylvania, Institute for Biomedical Informatics, Philadelphia, Pennsylvania, USA; Daniel J. Goldberg, Washington University in St. Louis, Department of Computer Science and Engineering, St. Louis, Missouri, USA; Natasha L. Ray, Princeton Day School, Princeton, New Jersey, USA; Praneel Chakraborty, University of Pennsylvania, Wharton School, Philadelphia, Pennsylvania, USA, University of Pennsylvania, School of Arts and Sciences, Philadelphia, Pennsylvania, USA; Daniel Himmelstein, Washington University in St. Louis, Denver, Colorado, USA, University of Pennsylvania, Department of Systems Pharmacology & Translational Therapeutics, Philadelphia, Pennsylvania, USA; Weixuan Fu, University of Pennsylvania, Institute for Biomedical Informatics, Philadelphia, Pennsylvania, USA; Jason H. Moore, University of Pennsylvania, Institute for Biomedical Informatics, Philadelphia, Pennsylvania, USA.

The original prototype release of PMLB (v0.2) [2] received positive feedback from the ML community, reflecting the pressing need for a collection of standardized datasets to evaluate models without intensive preprocessing and dataset curation. As the repository becomes more widely used, community members have requested new features such as additional information about the datasets, a standardized metadata schema, and new functions to find and select datasets given specific criteria, among others. In this Applications Note, we review PMLB's core functionality and present new enhancements that facilitate fluid interactions with the repository, both from the perspective of database contributors and end-users.

## 2 IMPLEMENTATION

PMLB consists of 3 main components: (1.) The collection of benchmark datasets, including metadata and associated documentation, (2.) a Python interface for easily accessing the datasets in the PMLB collection, and (3.) an R interface providing similar functionality to the Python interface. PMLB synthesizes and standardizes hundreds of publicly available datasets from diverse sources such as the UCI ML repository and OpenML, enabling systematic assessment of ML methods using a single data interface. Copies of the individual datasets are stored in the GitHub repository using Git Large File Storage, and each dataset is accompanied by a user-provided set of metadata describing the dataset (including keywords that can be used to categorize datasets), as well as an automatically generated Pandas Profiling report that quantitatively describes various characteristics of each dataset.

### 2.1 New datasets with rich metadata

Since PMLB's original release (v0.2) [2], we have made substantial improvements in collecting new datasets. PMLB now includes benchmark datasets for regression problems (in addition to classification problems, which have been supported since earlier versions). Each dataset now includes a `metadata.yaml` file containing general descriptive information, including the original web address of the dataset, a text description of its purpose, any associated publications, keywords, and descriptions of individual features and their coding schema, among others. Metadata files are supported by a standardized format that is formalized using JSON-Schema (version `draft-07`) [3]. Upcoming releases of PMLB improve upon the automated validation of datasets and metadata files to simplify contributions and maintain data accuracy.

### 2.2 User-friendly interfaces

The new version of PMLB includes a contribution guide with step-by-step instructions on how to add new datasets, edit existing datasets, or improve the Python or R interfaces. When a user adds a new dataset, summary statistics are automatically computed, a profiling report is generated (see below), a corresponding metadata template is created. Once changes are approved, PMLB's list of available datasets is automatically updated.

On PMLB's homepage, users can now browse, sort, filter, and search for datasets using a responsive table that includes summary statistics (**Figure 1**). In addition to the existing Python interface for PMLB, we have included an R library for interacting with PMLB. The R library includes a number of detailed "vignette" documents to help new users learn how to use the software. The website includes API reference guides detailing all user-facing functions and variables in PMLB's Python and R libraries.

### 2.3 Pandas profiling reports

We generate summary statistic reports for each dataset using `pandas-profiling`. These reports provide detailed quantitative descriptions of each dataset, including correlation structures between features and flagging of duplicate and missing values. Browsing the reports allows users and contributors to rapidly assess dataset quality and make any necessary changes. For example, if a feature is flagged as containing a single value repeated across all samples, it is likely that the feature is uninformative and should be removed from ML analyses. Profiling reports can be accessed either by navigating to the dataset's directory in the PMLB code repository, or by clicking the dataset name in the interactive dataset browser on the PMLB website.

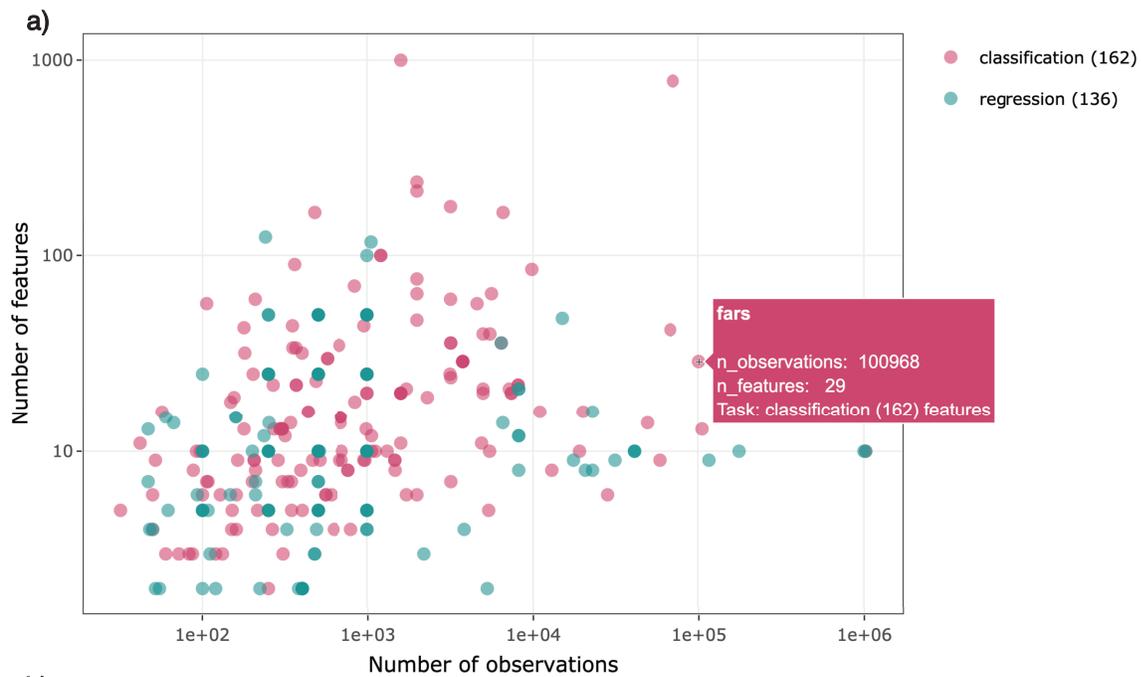

Fig. 1. Database search features on PMLB's website. **a)** Interactive scatterplot of databases in PMLB, showing number of features and number of observations in each dataset, as well as whether it is a regression or classification dataset. **b)** Responsive table of PMLB databases. Users can sort on any columns' values or filter based on ranges of values. Clicking on any dataset name will bring the user to the Pandas Profiling report for that dataset.

## 3 AVAILABILITY

PMLB is publicly available, open-source, and released under the MIT license. User-friendly interfaces are available for the Python and R programming languages, and can be installed via the Python Package Index (PyPI) and the Comprehensive R Archive Network (CRAN), respectively. The source code repository for PMLB is maintained at https://github.com/EpistasisLab/pmlb, and documentation for PMLB is provided at https://epistasislab.github.io/pmlb.

## ACKNOWLEDGEMENTS

We thank the open-source community for their valuable contributions and improvements made to PMLB during its development. We also especially thank GitHub user makeyourownmaker for original contributions to PMLB that were adapted into the interface for the R programming language.

## FUNDING

PMLB is developed with support from NIH grants `R01-AI116794`, `R01-LM010098`, `R01-LM012601` (PI: Jason Moore), `T32-ES019851` (PI: Trevor Penning), and `K99-LM012926` (PI: William La Cava). The authors have no relevant conflicts of interest.